\documentclass[letterpaper, 10 pt, conference]{ieeeconf}
\IEEEoverridecommandlockouts
\usepackage{amsmath,amssymb,amsfonts}
\usepackage{bm}
\usepackage{graphicx}
\usepackage{cite}
\usepackage{tikz}
\usepackage{booktabs}
\usepackage{xcolor}
\usepackage{capt-of}
\usepackage[hidelinks]{hyperref}
\usepackage{microtype}
\usetikzlibrary{positioning,arrows.meta,fit}

\newcommand{\vect}[1]{\bm{#1}}

\usepackage{titlesec}

\titlespacing*{\section}
{0pt}{1.2ex plus .5ex minus .2ex}{0.8ex plus .2ex}

\titlespacing*{\subsection}
{0pt}{0.85ex plus .3ex minus .2ex}{0.6ex plus .2ex}

\titlespacing*{\subsubsection}
{0pt}{0.8ex plus .2ex}{0.5ex}

\titlespacing*{\paragraph}
{0pt}{0.8ex}{0.5em}

\newif\ifanonymous
\anonymousfalse      

\begin{document}

\title{\LARGE \bf REFLEX: A Learned On-Board Reflex for Robust Autonomous Collision Recovery of Soft Quadrotors}

\ifanonymous
    \author{Author Names Omitted for Review}   
\else
     \author{%
    Jakob Hermeler$^{1,*}$,
    Luca Girardi$^{1,*}$,
    Gabriel Maquignaz$^{1}$,
    Stefan Leutenegger$^{1,\dagger}$,
    and Stefano Mintchev$^{2,\dagger}$%
    \thanks{$^{*}$Equal contribution. $^{\dagger}$Joint advising.}%
    \thanks{$^{1}$Mobile Robotics Lab,
    Department of Mechanical and Process Engineering,
    ETH Zurich, Zurich, Switzerland.
    \texttt{girardil@ethz.ch}}%
    \thanks{$^{2}$Environmental Robotics Lab, Faculty of Agricultural, 
    Nutritional and Engineering Sciences, University of Bonn, Bonn, Germany.}%
    \thanks{Supplementary video: \url{https://youtu.be/Q76GmW9cNgE}}%
    \thanks{Supported by EU Horizon Europe project SPEAR (grant no.~101119774).}%
    }%
\fi

\IEEEaftertitletext{%
    \vspace{-9pt}
    \begin{minipage}{\textwidth}
        \centering
        \includegraphics[
            width=\textwidth,
            trim=0 20 0 0,
            clip
        ]{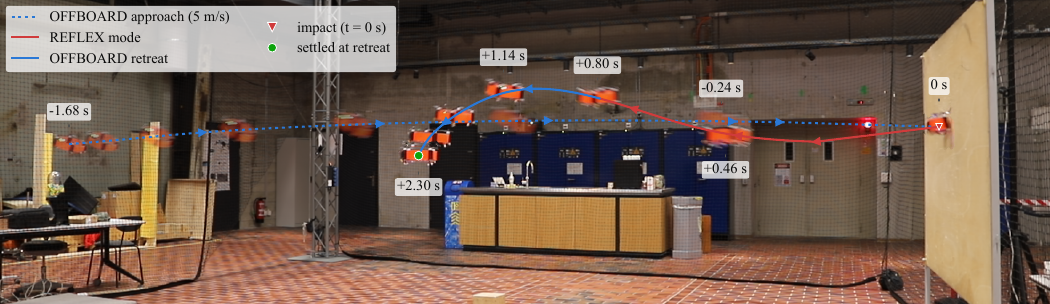}
        \captionof{figure}{%
            REFLEX recovering from a wall collision at 5\,m/s.
            The cycle shown is one of 15 consecutive recoveries
            within a single flight. Video frame composite sequence
            of the collision: approach (dotted); recovery from
            impact to the settled hold at the retreat point (solid).
        }
        \label{fig:hero}
        \vspace{4 pt}
    \end{minipage}%
}

\maketitle
\thispagestyle{empty}

\begin{abstract}
Collision-resilient airframes allow quadrotors to survive impacts, but do not ensure autonomous post-collision recovery. Innovation gating in stock onboard estimators often rejects valid external velocity measurements after high-speed impacts, compromising recovery. We present REFLEX, a collision-recovery flight mode that runs entirely on the flight controller, requiring neither contact sensors nor a companion computer. Its proprioceptive recurrent policy stabilizes the vehicle without position, velocity, heading, or altitude estimates. The translational estimate is reinitialized while REFLEX is active, and a trust gate checks estimator consistency before returning control to the stock autopilot. The policy is trained in simulation using proximal policy optimization, with episodes initialized from post-impact states measured in 47 real collisions. In hardware experiments, REFLEX recovered 19 of 22 wall impacts at 5.03 $\pm$ 0.09 m/s, compared with 1 of 14 for the matched stock autopilot. These included 15 consecutive recoveries in a single flight. A separate exploratory flight demonstrated two additional recoveries at 6.02 and 6.05\,m/s within the same flight, exceeding the previously reported velocity for autonomous wall-collision recovery by 64\%.
\end{abstract}

\section{Introduction}
\label{sec:intro}

Quadrotors are becoming autonomous tools for inspection, monitoring, and search and rescue~\cite{floreano_science_2015}, gaining particular relevance in cluttered environments such as forests, industrial interiors, and underground spaces~\cite{depetris2022ctmavsurvey,ren2022bubble}. These missions often require flight near obstacles in poor visibility to reach areas inaccessible to people or larger platforms.

Range requirements motivate faster flight~\cite{bauersfeld2022range}, which reduces the available time for obstacle detection and makes high-speed avoidance challenging~\cite{falanga2019howfast,ren2022bubble}. Glass and other transparent obstacles further limit avoidance because they are difficult to detect reliably~\cite{wang2024airbumper,mulgaonkar2020tiercel}. Impact recovery must therefore complement avoidance as a fallback layer.

Recovery requires both structural survival and restoration of controlled flight. Collision-tolerant designs address the former, but the latter remains challenging~\cite{depetris2022ctmavsurvey} (Sec.~\ref{sec:related}). FlexiQuad, used as the experimental platform in this work, is a soft collision-resilient quadrotor whose airframe survives frontal wall impacts without structural or functional damage~\cite{girardi2025flexiquad}; nevertheless, the stock autopilot does not reliably restore controlled flight.(Sec.~\ref{subsec:m-hardware}).

At high-speed impact, the vehicle's velocity reverses within tens of milliseconds, and the inertial sensors briefly saturate. In our experiments with stock PX4, a commonly-used open-source autopilot, the estimator's innovation gate rejects valid external velocity measurements after impact. The filter retains a near-zero velocity estimate while the vehicle rebounds and drifts backwards, leaving the controller to track its setpoint using an erroneous translational state. Rejection worsens with impact speed because the velocity jump increases while the gate threshold remains fixed. Consequently, the same unmodified autopilot recovers most trials at 3\,m/s but almost none at 5\,m/s (Sec.~\ref{sec:experiments}).
Recovery must begin while the translational estimate is still unusable. 

We propose REFLEX, a PX4 collision-recovery mode that runs entirely on the flight controller, requiring neither contact sensors nor a companion computer. Its learned proprioceptive recurrent policy stabilizes the vehicle using accelerometer specific force, gyroscope body rates, and body-frame gravity direction. As these signals undergo bounded transient disturbances during impact, the policy can stabilize the vehicle from them alone. It does so by commanding the stock rate loop directly, bypassing all control layers that depend on the translational estimate. After a fixed delay from engagement, REFLEX reinitializes that estimate, and a trust gate checks consistency with incoming measurements before returning control to the stock autopilot.

Our contributions are threefold. \textbf{(1)} We experimentally characterize autonomous post-impact recovery on a highly compliant, collision-resilient quadrotor adapted from FlexiQuad~\cite{girardi2025flexiquad}, linking failures of stock PX4 to rejection of valid post-impact velocity measurements. \textbf{(2)} We introduce REFLEX, a recovery architecture for this soft platform that combines proprioceptive recurrent stabilization, estimator reinitialization, and consistency-gated handback, implemented entirely on the flight controller. The policy is trained in simulation using proximal policy optimization (PPO) with an asymmetric actor--critic architecture. Training uses post-impact states measured in 47 real collisions, and the learned policy transfers zero-shot to hardware despite the large attitude excursions, transient sensor saturation, and structural oscillations observed after impact. \textbf{(3)} We validate autonomous recovery through comparative hardware experiments at 3 and 5\,m/s. REFLEX recovers from 19 of 22 impacts (86\%) at $5.03 \pm 0.09$\,m/s, including 15 consecutive recoveries in one flight, compared with 1 of 14 (7\%) for the matched stock autopilot. At $3.22 \pm 0.05$\,m/s, REFLEX recovers from 9 of 11 impacts (82\%). A separate exploratory flight demonstrated recovery in two of two attempts, at 6.02 and 6.05 \, m/s. The code and raw experimental data will be made publicly available upon acceptance.

\section{Related Work}
\label{sec:related}

\subsection{Surviving and Recovering from Collisions}
\label{subsec:rw-surviving}

Protective cages~\cite{briod2014gimball}, inflatable bodies~\cite{nguyen2023sobar}, and compliant frames~\cite{depetris2025morphy,girardi2025flexiquad,pham2026holoarm} improve collision tolerance at a structural level. Several platforms also demonstrate post-impact flight recovery. Morphy restores hover after midflight collisions at up to 3\,m/s using sensorized compliant arms and deflection-aware control allocation~\cite{depetris2025morphy}. HoLoArm combines passively deformable arms with an RL policy to re-stabilize after frontal impacts, without reporting impact speed~\cite{pham2026holoarm}. A quadrotor equipped with deformable Tombo propellers arrests its post-impact fall when the position reference is shifted away from the obstacle~\cite{bui2023tombo}.

Studies focused on recovery report autonomous wall-collision recovery at 2.0\,m/s~\cite{dicker2017quadrotor}, 2.58\,m/s~\cite{liu2021impact}, 3.5\,m/s~\cite{liu2023dynamic}, and 3.7\,m/s~\cite{bredenbeck2025tactile}. These platforms use different mechanical strategies to tolerate impact. Dicker et al.\ employ propeller-protective bumpers~\cite{dicker2017quadrotor}, Liu and Karydis use compliant arms with passive shock absorbers~\cite{liu2021impact, liu2023dynamic}, and Bredenbeck et al.\ use an icosahedral tensegrity shell~\cite{bredenbeck2025tactile}. The latter two approaches also sense contact through Hall-effect measurements of arm deformation~\cite{liu2023dynamic} or binary contact sensors at the tensegrity vertices~\cite{bredenbeck2025tactile}. Their recovery controllers depend on valid, externally aided state estimates, a limitation explicitly identified by Liu and Karydis~\cite{liu2023dynamic}. They also demonstrate the effect of airframe compliance: under the same controller and recovery strategy, a rigid frame fails on every trial at 3.5\,m/s, whereas a compliant frame recovers on all trials~\cite{liu2023dynamic}.

Higher-speed impacts have been demonstrated only outside autonomous in-flight recovery. Battiston et al.\ report piloted wall impacts of up to approximately 4 \, m/s~\cite{battiston2019attitude}. FlexiQuad survives frontal impacts at 4.5\,m/s without structural or functional damage, but autonomous post-impact recovery is not addressed~\cite{girardi2025flexiquad}. Zha et al.\ demonstrate a tensegrity vehicle surviving a 7.8\,m/s wall impact, but flight resumes only after landing, reorienting, and taking off again~\cite{zha2024tensegrity}.

\subsection{Learned Control on Flight Hardware}
\label{subsec:rw-learned}

Learned policies have demonstrated recovery in several settings. Hwangbo et al.\ stabilize hand-thrown quadrotors using a general control policy~\cite{hwangbo2017control}, while Zhao et al.\ recover quadrotors from arbitrary ground-resting attitudes using a recurrent policy without state-estimate inputs~\cite{zhao2026agilefall}. Both policies run on companion computers.

Embedded neural control has also been demonstrated directly on flight controllers: a 5000-parameter policy runs at 500\,Hz on Crazyflie~\cite{molchanov2019sim}; RAPTOR deploys recurrent control across PX4, Betaflight, Crazyflie, and M5StampFly~\cite{eschmann2026raptor}; and a neural flight mode runs at 650\,Hz on Pixracer Pro~\cite{hegre2025neuralmode}. HoLoArm likewise executes its RL policy directly on a PX4 flight controller~\cite{pham2026holoarm}. These embedded policies use position and velocity estimates, the same states that are corrupted by higher-speed collisions.

Action-space comparisons report more reliable transfer under model mismatch with policies commanding collective-thrust and body-rate (CTBR) than those commanding individual rotor-thrust commands~\cite{kaufmann2022benchmark}, with similar findings for fall recovery~\cite{zhao2026agilefall}. REFLEX adopts this action space while excluding position, velocity, heading, and altitude estimates from its recurrent stabilization policy (Sec.~\ref{subsec:m-policy}). To our knowledge, no collision recovery policy has been demonstrated to run on the flight controller and remain operational while those estimates are invalid.

\subsection{State Estimation and Recovery}
\label{subsec:rw-estimator}

Recovery control depends on which state estimates remain usable during impact and how corrupted or unavailable estimates are restored. Battiston et al.\ investigate collision-induced inertial sensor disturbances and compare conventional, robust, and adaptive attitude estimators using gyroscope, accelerometer, and magnetometer measurements~\cite{battiston2019attitude}. Their alternatives to the stock estimator gate accelerometer use during large linear accelerations. In closed-loop tests, a multiplicative extended Kalman filter (EKF) and an adaptive $H_\infty$ filter produce faster and more consistent attitude recovery than the tested PX4 complementary filter, although adaptation provides limited additional benefit and requires careful tuning. Their recovery sequence stabilizes attitude and height before returning control to the pilot; restoration of autonomous position control is not demonstrated.

Bredenbeck et al.\ explicitly incorporate collisions into Kalman-filter prediction~\cite{bredenbeck2025tactile}. Binary tactile contacts and pre-impact linear and angular velocities drive an impulse model that predicts post-impact
velocities and rates. This addresses the delayed response of a collision-agnostic filter, which their simulations show can
produce incorrect recovery commands. The resulting estimate supports a position-based retreat and subsequent path
adjustment to avoid repeated contact. Their prototype runs the motion-capture-assisted estimator and high-level control offboard, with contact sensing and rate control onboard.

Faessler et al.\ address unavailable visual state estimates through staged recovery~\cite{faessler2015automatic}. The vehicle first stabilizes attitude using inertial measurements, then vertical motion using a downward range sensor, before reinitializing visual-inertial estimation and progressively restoring velocity and position control. Stage transitions depend on vehicle-state thresholds and estimator availability. This onboard system uses a companion computer and is demonstrated through free-space throws rather than collisions.

REFLEX extends staged estimator recovery to high-speed collisions in which the tested PX4 estimator rejects valid external velocity measurements. Its contribution is a recovery mode implemented entirely on the flight controller that tolerates the resulting loss of usable translational feedback and restores autonomous position control, without contact sensing or an explicit impact model.

\section{Methodology}
\label{sec:methodology}

\begin{figure*}[hbt]
  \centering
  \includegraphics[width=\textwidth,trim=5 8.5 5 6,clip]{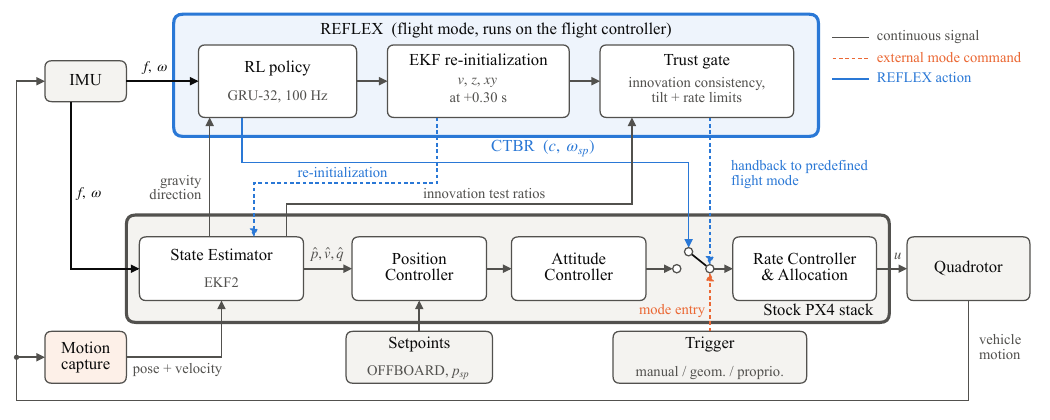}
  \caption{REFLEX (blue) within the stock PX4 flight stack (black). The policy commands collective thrust and body rates (CTBR) at 100\,Hz, bypassing position and attitude control. Translational reinitialization begins 0.30\,s after engagement, followed by consistency-gated handback.}
  \label{fig:system}
\end{figure*}

\subsection{Problem Formulation and Recovery Objective}
\label{subsec:m-problem}

We consider autonomous recovery of a collision-resilient quadrotor after a frontal impact at several meters per second that leaves the vehicle airborne and its propulsion system operational. Nominal position control uses an onboard estimator that fuses inertial measurements with external aiding. We address the regime in which the translational estimate becomes unsuitable for feedback, while inertial measurements and the estimated gravity direction remain sufficiently informative for short-term stabilization despite transient impact disturbances. The experimental basis for this operating regime is presented in Sec.~\ref{subsec:e-baseline}.

Given a recovery trigger, the objective is to regain controlled flight and resume nominal position control without landing or pilot intervention. The design targets are \textbf{(R1)} stabilization without translational estimates while limiting post-impact drift; \textbf{(R2)} restoration of usable translational estimation; and \textbf{(R3)} assessment of estimator consistency and vehicle stabilization before nominal handback. Recovery computation must run entirely on the flight controller. 

\subsection{Recovery Architecture}
\label{subsec:m-overview}

REFLEX is registered as a flight mode in PX4. On engagement, it zeroes the recurrent policy's hidden state, arms the estimator-reinitialization timer, and takes control of the stock rate-controller setpoints (Fig.~\ref{fig:system}). The policy stabilizes the vehicle without translational estimates (R1). While it remains active, the module reinitializes the translational estimate (R2), then checks estimator consistency and vehicle stabilization before nominal handback to a preselected stock mode (R3). All experiments use OFFBOARD as the handback mode. The timeout and pilot-input exceptions are specified in Sec.~\ref{subsec:m-repair}.

Policy inference runs at 100\,Hz, and all recovery computation executes on the flight controller without a companion computer (Sec.~\ref{subsec:e-cost}). The module responds to mode entry rather than a particular trigger source; manual, position-based, or proprioceptive collision detection~\cite{tomic2017external} can provide the entry command. The experiments use an external position-based trigger (Sec.~\ref{subsec:e-setup}). Motion-capture measurements support that trigger, estimator reinitialization, and the handback check, but do not directly enter the policy observation. Alternative aiding and trigger sources remain possible, but are not evaluated in this work.

\subsection{Proprioceptive RL Policy}
\label{subsec:m-policy}

The policy receives a 13-dimensional observation comprising body-frame specific force, filtered body rates, estimated body-frame gravity direction, and the previous action. Specific force is $\vect{f}=\vect{a}-\vect{g}$, where $\vect{a}$ is kinematic acceleration and $\vect{g}$ is gravitational acceleration, expressed in the same frame. Position, velocity, heading, and altitude estimates are excluded to avoid relying on corrupted estimates. The only estimator-derived input is the gravity direction, extracted from the third row of the attitude rotation matrix. It depends on roll and pitch but is invariant to yaw. The filter propagates attitude using gyroscope measurements; brief saturation at impact can perturb this propagation, with the resulting gravity-direction error quantified in Sec.~\ref{subsec:e-baseline}.

Specific force and body rates pass through the stock second-order low-pass filters at 30 and 25\,Hz, respectively, without additional filtering. These filters attenuate the brief saturation peaks in the policy inputs. Training reproduces the sensor limits and filtering of the stock system to present the policy with realistic saturated sensor data (Sec.~\ref{subsec:m-sim}).

The action comprises collective thrust and three body-rate commands. The stock rate controller tracks the commanded rates and forwards thrust to control allocation, bypassing the position and attitude controllers. This action space follows evidence of more reliable hardware transfer than direct rotor-thrust commands under model mismatch~\cite{kaufmann2022benchmark}. Compiled-in roll, pitch, and yaw rate limits are $220/220/200^\circ$/s, matching the stock PX4 authority. The yaw channel is masked and was not trained.

The network comprises Dense($13\rightarrow32$, tanh), a single-layer GRU-32, and Dense($32\rightarrow4$, linear), with 6916 parameters. The recurrent design follows the fall-recovery policy of~\cite{zhao2026agilefall}; embedded GRU inference has also been demonstrated on flight controllers~\cite{eschmann2026raptor}. The hidden width of 32 was selected to fit the flash budget.

\subsection{Estimator Reinitialization and Handback}
\label{subsec:m-repair}

To restore the feedback required by nominal position control, REFLEX reinitializes velocity, height, and horizontal position together in a single filter update, 0.30\,s after engagement, while the policy remains active. All three states use external-vision aiding, the only exteroceptive source in our experiments; samples older than 0.50\,s are rejected, leaving the corresponding state unchanged. For velocity, a live aiding sample is preferred over the last finite external-vision sample stored in a cache. If neither is available, velocity is reset to zero with a standard deviation of 1.5\,m/s, selected to avoid triggering the velocity failsafe. Attitude is deliberately excluded from reinitialization to keep the gravity-direction input used by the policy continuous.

Nominal handback requires a completed reinitialization, valid local velocity, a settled vehicle, and a finite velocity innovation test ratio below 2.5. PX4's \texttt{vel\_test\_ratio} is the largest filtered innovation ratio of the external-vision velocity measurements across the three axes; a value of 1.0 corresponds to the estimator's own rejection threshold. The vehicle is considered settled when tilt is below $20^\circ$ and the roll--pitch rate norm is below 3.0\,rad/s. All conditions must hold continuously for 0.1\,s within a window from 0.5 to 1.5\,s after engagement. If the window expires without nominal handback, REFLEX requests the preselected fallback mode. Pilot stick input defers both handback and timeout. No explicit velocity-magnitude bound is imposed because the policy does not observe translational drift.

The gate evaluates velocity-innovation consistency across all three axes, including vertical velocity, but does not check height consistency and is disabled after handback. While REFLEX remains active, a velocity innovation test ratio exceeding 2.5 continuously for 0.3\,s triggers joint reinitialization of velocity, height, and horizontal position. This reinitialization can be repeated at most three times.

\subsection{Simulation and Policy Training}
\label{subsec:m-sim}

FlexiQuad is modeled in NVIDIA Isaac Sim as 24 rigid bodies connected by 48 D6 spring--damper joints in NVIDIA PhysX. Stiffness and damping are fitted to reproduce the experimentally validated finite-element model~\cite{girardi2025flexiquad} under quasi-static squeezing, out-of-plane bending, and dynamic compression at wall collision. The simulator includes ports of the PX4 v1.17 rate controller and mixer with flight gains matching those used in real experiments, a synthetic IMU with the stock low-pass and notch filters, measurement limits, actuator transport delay, and motor lag.

The collision itself is not simulated. Controlled real collisions provide motion-capture trajectories and onboard records of inertial measurements and actuator commands. Recorded flight data are used to identify the actuator and sensor models, namely motor spin-up and spin-down time constants of 31 and 20\,ms and a 3 to 9\,ms actuator delay from electronic-speed-controller telemetry against motor commands, linear drag, and inertial-measurement noise levels. The same records provide post-impact initial states for policy training. Episodes' initial states comprise position, attitude, and linear and angular velocity, sampled from 30 to 300\,ms after real impacts in 10\,ms steps. The resulting dataset contains 1306 states from 47 collisions disjoint from all evaluation trials reported in Sec.~\ref{sec:experiments}.

Seventy percent of training episodes are initialized from uniformly sampled measured post-impact states. The remaining 30\% use synthetic states spanning tilts up to $120^\circ$ and angular rates up to $2500\,{}^\circ/\mathrm{s}$, beyond the gyroscope limit of $2000\,{}^\circ/\mathrm{s}$. Randomization includes sensor noise, synthetic excitation of frame oscillations, physical parameters, and injected attitude errors exceeding those measured in flight (Sec.~\ref{subsec:e-baseline}).

The policy is trained to approach a level, low-speed hover. At policy step $t$, the reward is
\begin{align}
  r_t ={}& 2.0 - 3.0\,\theta - 0.1\,\lVert\omega\rVert
    + 1.5\,\varphi(v_z) + 1.0\,\varphi(v_{xy}) \nonumber\\
    & - 0.05\,\lVert\Delta a_t\rVert_2
    - 0.1\,(a_{t,0} - a_{\mathrm{hover}})^2
    + 1.0\,\mathbb{I}_{\mathrm{set}},
  \label{eq:reward}\\
  R_t ={}& r_t - 50\,\mathbb{I}[\text{crashed} \lor \text{ceiling exit}],
  \label{eq:terminal}
\end{align}
where $\theta$ is tilt, $\omega$ denotes body rates, $v_z$ is vertical velocity, $v_{xy}$ is horizontal speed, and $a_t$ is the action. Its thrust component is $a_{t,0}$, with hover value $a_{\mathrm{hover}}$, and $\Delta a_t=a_t-a_{t-1}$. The velocity terms use the bounded reward function $\varphi(v)=1/[1+(v/s)^2]$, which is maximal at zero velocity and falls to half its maximum at $|v|=s=0.5$\,m/s. $\mathbb{I}[\cdot]$ is the indicator function, and
\[
\mathbb{I}_{\mathrm{set}}=
\mathbb{I}[\theta<25^\circ\land\lVert\omega\rVert<90^\circ/\mathrm{s}
\land |v_z|<0.5\,\mathrm{m/s}].
\]
The learner receives $R_t$; the crash or ceiling-exit penalty is applied once at termination. Attitude, rate, and velocity terms favor a level, low-speed state, while action-change and hover-thrust penalties regularize the commands. The in-set bonus applies at each qualifying step; the simulation success criterion requires the same set to be maintained for 0.2\,s.

Training uses PPO with a recurrent actor and a non-recurrent critic~\cite{zhao2026agilefall}. The actor receives the policy observations defined in Sec.~\ref{subsec:m-policy}. The critic additionally receives true kinematics, injected attitude-error magnitude, and the episode's randomization parameters. Only the actor is deployed on the flight controller.

\section{Experimental Evaluation}
\label{sec:experiments}

The experiments characterize post-impact estimation errors on real hardware in the tested stock PX4 configuration and compare autonomous recovery with and without REFLEX under matched wall-impact conditions. We also assess the spatial requirements of recovery and the computational cost on the flight controller.

\subsection{Hardware Platform and Protocol}
\label{subsec:e-setup}
\label{subsec:m-hardware}

\begin{figure}[t]
  \centering
  \includegraphics[width=\columnwidth,trim=0 5 0 3,clip]{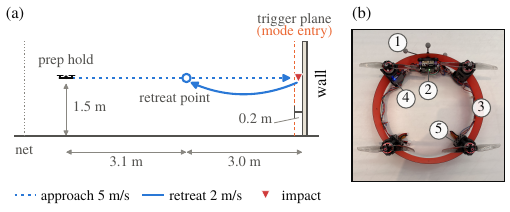}
  \caption{Platform and protocol. (a) Side schematic of the 5\,m/s experiment, showing the prep hold, trigger plane, wall, and commanded retreat point. (b) Top view of the 0.371\,kg vehicle (front pointing downward): (1) rear-mounted motion-capture markers, (2) flight controller, (3) flexible ring, (4) WiFi bridge, and (5) motor hub.}
  \label{fig:setup}
\end{figure}

The vehicle is based on FlexiQuad~\cite{girardi2025flexiquad}, whose soft ring-shaped frame is approximately 1000 times more compliant than its rigid counterparts and deforms elastically to survive frontal wall impacts without structural or functional damage. Our adapted vehicle weighs 0.371\,kg including its four batteries and has a 216\,mm-diameter ring made from woven glass-fiber composite strips (FR-4 HF), with motor hubs printed in Tough 2000 resin. A Pixracer Pro flight controller runs PX4 v1.17, with accelerometer and gyroscope measurement limits of $\pm16$\,g and $\pm2000^\circ$/s, respectively.

Experiments take place in an indoor flight space with a rigid wooden panel supported by two metal stands. A Qualisys motion-capture system supplies both ground truth and the external measurements used by the onboard estimator. EKF2 fuses inertial measurements with motion-capture-derived position, velocity, and yaw received through PX4's external-vision interface. 

A WiFi MAVLink bridge carries commands, telemetry, and motion-capture data, including the external mode-entry command and OFFBOARD setpoints. The stabilization policy runs locally without streamed control commands; external aiding remains involved in estimator reinitialization and the handback check (Sec.~\ref{subsec:m-overview}).

For each trial, the offboard controller first holds the vehicle at the prep position at an approach height of 1.5\,m (Fig.~\ref{fig:setup}(a)). On the operator's command, it initiates the approach at the prescribed speed, with the position setpoint 2.5\,m behind the wall to prevent pre-impact braking. The mode-entry command is issued when the tracked marker centroid crosses a plane 0.2\,m before the wall. Because the centroid lies behind the front of the ring, this occurs just after first contact and before the main deceleration. After stabilization and handback to OFFBOARD, the controller commands a retreat to a point 3\,m from the wall at 2\,m/s. The next trial starts from that position.

Baseline trials follow the same approach protocol without REFLEX: the retreat setpoint is issued at impact, and the stock controller remains in OFFBOARD throughout. The approach is configured to avoid impact-induced overturning. Mean pre-impact pitch toward the wall ranges from approximately $10^\circ$ to $13^\circ$ across test groups (Table~\ref{tab:outcomes}). The approach targets a forward pitch below $15^\circ$, motivated by impact-induced overturning observed at arrival pitch above $20^\circ$ in this setup. Measured pre-impact pitch is reported in Table~\ref{tab:outcomes}.

The comparative evaluation contains 22 REFLEX and 14 baseline impacts at a commanded approach speed of 5\,m/s. A lower-speed tier contains 11 REFLEX and 15 baseline impacts at 3\,m/s, using a retreat point 1\,m from the wall and a retreat speed of 1\,m/s. Additional 6\,m/s demonstrations are reported separately in Sec.~\ref{subsec:e-results}.

A trial succeeds if OFFBOARD holds authority at completion and the vehicle reaches the commanded retreat position within 10\,s of impact and holds it for 1.0\,s. REFLEX must therefore enter recovery mode and return to OFFBOARD; the baseline must remain in OFFBOARD. Speed and position metrics are computed from motion capture rather than the potentially corrupted onboard estimate.

\begin{table}[t]
  \caption{Collision outcomes and measured impact conditions.}
  \label{tab:outcomes}
  \centering
  \setlength{\tabcolsep}{3pt}%
  \scriptsize
  \begin{tabular}{lccccc}
\toprule
Experiment & $n$ & \% Rec. (abs) & Speed (m/s) & Pitch$^{\dagger}$ ($^\circ$) & Decel.$^{\ddagger}$ (m/s$^2$) \\
\midrule
REFLEX 5 m/s   & 22 & \textbf{86\%} (19) & $5.03\pm0.09$ & $9.9\pm2.4$  & $305\pm32$ \\
REFLEX 3 m/s   & 11 & \textbf{82\%} (9)  & $3.22\pm0.05$ & $11.7\pm0.5$ & $151\pm18$ \\
Baseline 5 m/s & 14 & 7\% (1)  & $5.10\pm0.10$ & $12.9\pm1.8$ & $326\pm42$ \\
Baseline 3 m/s & 15 & 67\% (10) & $3.21\pm0.13$ & $11.7\pm1.4$ & $149\pm15$ \\
\bottomrule
  \end{tabular}
  \par\smallskip
    \parbox{\columnwidth}{\raggedright\footnotesize Measured quantities from motion capture, reported as mean $\pm$ SD. 
    $^\dagger$Per-trial median pitch over the final 0.2\,s before impact.
    $^\ddagger$Peak wall-normal deceleration within $\pm0.05$\,s of contact..}
\end{table}

\subsection{Post-Impact Estimation and Baseline Behavior}
\label{subsec:e-baseline}

At 5\,m/s, impacts reverse the vehicle's velocity within tens of milliseconds and briefly saturate both inertial sensors. The gyroscope remains saturated for approximately 10\,ms and the accelerometer for approximately 60\,ms in short bursts, after which both resume unsaturated measurements. Across the characterized post-impact states, tilt has a median of $33^\circ$ and a maximum of $76^\circ$. Frame oscillations at 21\,Hz persist for approximately one second.

The stock stack recovers 1 of 14 (7\%) trials at 5\,m/s and 10 of 15 (67\%) at 3\,m/s with the same configuration. The motion-capture feed captures the impact-induced velocity reversal, and the measured decelerations are comparable between REFLEX and baseline at each speed. Although the 5\,m/s baseline approaches with a slightly greater pitch, it remains below the overturning regime targeted by the protocol (Table~\ref{tab:outcomes}).

\begin{figure}[t]
  \centering
  \includegraphics[width=\columnwidth,trim=2 9 2 6,clip]{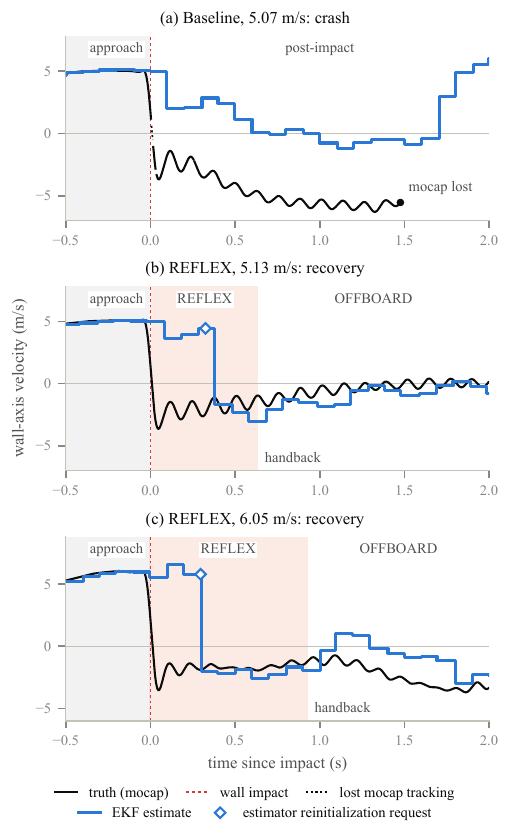}
\caption{Estimated and motion-capture-derived wall-axis velocity for (a) stock PX4 failure at 5.07\,m/s, (b) REFLEX recovery at 5.13\,m/s, and (c) exploratory recovery at 6.05\,m/s. Diamonds mark estimator-reinitialization requests. Mode annotations indicate REFLEX engagement and handback to OFFBOARD.}

  \label{fig:belieftruth}
\end{figure}

The logs show rejection of valid external velocity measurements by the estimator's innovation gate. In the representative 5\,m/s baseline trial of Fig.~\ref{fig:belieftruth}(a), the vehicle rebounds at more than 4\,m/s while its estimated velocity remains near zero. During rejection, the EKF propagates velocity using inertial measurements without external-velocity corrections. Every 5\,m/s baseline trial rejects the feed for at least 1\,s after impact, and several reject it for up to 2\,s. The controller consequently tracks its retreat setpoint using erroneous translational feedback.

At 3\,m/s, the median rejection duration is also approximately 1\,s, but velocity errors are smaller: the median is 0.6\,m/s. In 33\% of these trials, the estimator resumes accepting the feed at the flown gate setting. At 5\,m/s, analysis of the recorded innovations indicates that gate settings a median of 3.4 times the flown value (and up to ten times) would be needed to accept these measurements. Nevertheless, increasing the threshold would risk accepting motion-capture glitches observed during normal flight, where an accepted corrupted sample can produce meter-scale estimation jumps.

Measurement acceptance alone does not ensure rapid reconvergence in the tested configuration. In the two 5\,m/s baseline trials where reversal measurements were accepted, the velocity estimate nevertheless failed to reconverge within the first 1.5\,s after impact, and median velocity errors remained above 4\,m/s, as in all 14 baseline trials. The observed covariance-weighted updates corrected the large error too slowly for recovery within the available space. These observations motivate the discrete reinitialization in Sec.~\ref{subsec:m-repair}.

The rotational channel remains substantially more accurate. Within the first second, the median gravity-direction error is $5^\circ$, compared to below $2^\circ$ before impact, without sustained tilt drift. Gyroscope rates agree with motion-capture-derived rates over the same interval: the reported medians are 210 and 211$^\circ$/s, respectively, with per-axis correlation around 0.9. The height estimate remains within decimeters of ground truth, while velocity errors reach several meters per second. This separation supports stabilization using inertial measurements and estimated gravity direction despite corrupted translational feedback.

Ten baseline trials reach the safety net at 7.5\,m from the wall without controlled recovery. The two trials that accepted reversal measurements reach wall distances of 7.26 and 7.32\,m, but still fail to recover; even the successful trial reaches 6.0\,m despite a 3\,m retreat command. While a larger arena might allow additional recoveries,the excursion in the single successful trial requires substantial obstacle-free space.

\subsection{REFLEX Recovery Performance}
\label{subsec:e-results}

REFLEX recovered from 19 of 22 wall impacts (86\%) at $5.03\pm0.09$\,m/s (mean $\pm$ standard deviation), compared with 1 of 14 (7\%) for the matched stock autopilot (Table~\ref{tab:outcomes}). All 22 trials used the same vehicle setup. The final 15 trials were all successful and occurred in a single uninterrupted flight, at impact speeds of 4.89--5.13\,m/s and intervals of approximately 21\,s, without landing, flight-controller restart, hardware repair, or pilot intervention during recovery. Fig.~\ref{fig:hero} shows one complete recovery cycle. Across the 19 successful trials, median times from impact to handback and settling at the retreat position were 0.74 and 2.5\,s, respectively.

\begin{figure}[t]
  \centering
  \includegraphics[width=\columnwidth,trim=5 10 5 6,clip]{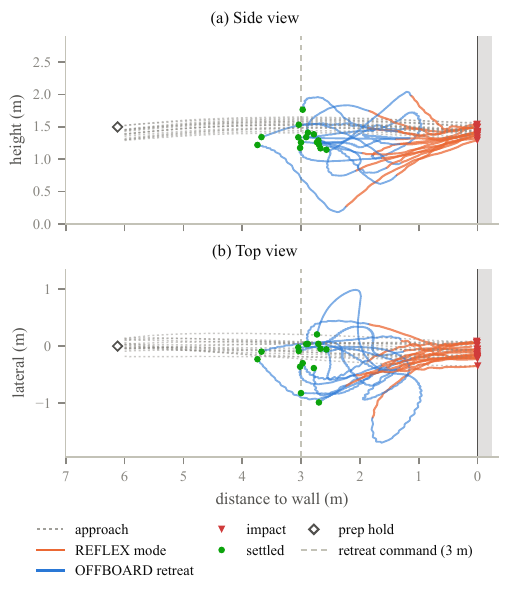}
  \caption{Fifteen consecutive recoveries at 4.89--5.13\,m/s in one flight. Motion-capture trajectories in (a) side and (b) top view show approach (dotted), REFLEX control (orange), OFFBOARD retreat (blue), and settled positions (green) near the 3\,m retreat command.}
  \label{fig:streak}
\end{figure}

While REFLEX is active, the policy has no position or velocity feedback, so translational drift determines the space required for stabilization. In the 15-recovery flight, all policy-controlled segments occupy a $2.2\times1.4\times1.5$\,m region in front of the wall despite drift along all three axes (Fig.~\ref{fig:streak}). Across these trials, the per-trial minimum post-impact height was $1.07\pm0.32$\,m (mean $\pm$ SD), with a lowest observed floor clearance of 0.3\,m (Fig.~\ref{fig:streak}(a)). After handback, each trajectory converges near the commanded retreat position 3\,m from the wall.

Table~\ref{tab:kpis} reports velocity-estimation error over 0.05--1.5\,s after impact: 0.75\,m/s for REFLEX versus the baseline values of 5.2\,m/s. Fig.~\ref{fig:belieftruth}(b) illustrates the successful sequence. The estimate initially remains near zero while the vehicle rebounds. The policy stabilizes and levels the vehicle's attitude using inertial measurements and estimated gravity direction. Reinitialization brings the velocity estimate close to motion capture, after which it tracks the measured motion through handback and retreat.

All three failures at 5\,m/s occurred after handback to OFFBOARD despite passing the configured checks. In the first, the reinitialized velocity estimate was within 0.3\,m/s of motion capture at handback, although rejection of external velocity measurements had already resumed. The estimate subsequently indicated motion in the wrong direction, and the pilot intervened. In the second, the autopilot switched to another estimator instance that reported a similar velocity-sign error, causing the commanded retreat to drive the vehicle back toward the wall. In the third, the gate missed a vertical-velocity error: the estimator indicated a slight descent while the vehicle climbed at 2.5\,m/s. The first two cases show that passing the handback checks does not ensure continued estimation accuracy; the third shows that an existing vertical error can pass undetected. These findings motivate continued estimator monitoring after handback and an explicit vertical-state consistency check (Sec.~\ref{sec:limitations}).

At 3\,m/s, REFLEX recovers 9 of 11 impacts (82\%), including eight consecutive recoveries in one flight at a 16\,s cadence, compared with 10 of 15 (67\%) for the stock stack. The difference in observed outcomes is smaller at this speed than at 5\,m/s, consistent with the speed-dependent baseline behavior in Sec.~\ref{subsec:e-baseline}.

 At 5\,m/s, the median maximum wall distance from impact to settling was 2.5\,m (range 2.5--3.2\,m) for the analyzed REFLEX recoveries. Only one of the 14 baseline trials settled, reaching a maximum wall distance of 6.0\,m. Ten of the 13 unsuccessful trials ended in the safety net, which physically limited their excursions. The median full-episode bounding volume was 14.5\,m$^3$ for the analyzed REFLEX recoveries and at least 164\,m$^3$ across all baseline trials, including failures (Table~\ref{tab:kpis}).

\begin{table}[t]
  \caption{Recovery metrics at 5\,m/s.}
  \label{tab:kpis}
  \centering
  \footnotesize
  \setlength{\tabcolsep}{6pt}
  \begin{tabular}{@{}lcc@{}}
    \toprule
    Metric & REFLEX & Baseline \\
    \midrule
    Median velocity error (m/s) & 0.75 & 5.2 \\
    Max. wall distance (m) & 2.5 & 6.0$^{\dagger}$ \\
    Median full-episode volume (m$^3$) & 14.5 & $\geq164^{\ddagger}$ \\
    \bottomrule
  \end{tabular}
  \par\smallskip
  \parbox{\columnwidth}{\raggedright\footnotesize
    Velocity error is measured against motion capture over 0.05--1.5\,s after impact ($n=22/14$ REFLEX/baseline). Wall distance is the maximum from impact to settling, reported as the median over 19 REFLEX trials. $^{\dagger}$Single settled baseline trial. Volume is the full-episode bounding-box volume, summarized over 19 successful REFLEX trials and all 14 baseline trials. $^{\ddagger}$Baseline excursions were limited by capture in the safety net in 10 of the 14 trials.}
\end{table}

A separate exploratory flight demonstrated successful recovery in two of two attempts, at impact speeds of 6.02 and 6.05\,m/s. The latter recovery is shown in Fig.~\ref{fig:belieftruth}(c). Recovery envelopes at these speeds were not systematically characterized, and these trials are not included in the comparative statistics.

\subsection{Computational Cost}
\label{subsec:e-cost}

A REFLEX cycle takes 6.76\,$\mu$s on average and 642\,$\mu$s at maximum, the latter below 7\% of the 10\,ms period of the 100\,Hz policy loop. In the same measurement, a stock rate-controller cycle takes 8.43\,$\mu$s and an EKF2 update 181--210\,$\mu$s. The work queue hosting REFLEX uses 2.7\% of the CPU; because it also hosts other modules, this is an upper bound on REFLEX's share. The module therefore fits within the flight controller's measured computational budget.

\section{Limitations and Future Work}
\label{sec:limitations}

The evaluation compares the integrated REFLEX system with the tested stock PX4 configuration. Improving the estimator's response to sensor saturation and abrupt velocity changes is a complementary direction that could improve recovery under both stock control and REFLEX. The learned policy provides short-term stabilization while translational feedback is unreliable, and improved estimation could reduce the failures observed after handback. The performance achievable through estimator adaptations alone, and the benefit of the learned policy relative to a classical recovery controller using the same reinitialization and handback logic, remain to be evaluated.

The handback gate checks estimator consistency over a short interval before release but does not continue monitoring afterwards. It also lacks a separate vertical-state consistency check, which led to one of the failure modes observed at 5\,m/s. Extending the recovery-mode re-divergence detector beyond handback and adding vertical-state checks are therefore priorities. The timeout requests the fallback mode even when the nominal gate has not passed; it is not a guarantee that the estimate is suitable for position control.

Position and height can drift while the policy is active because it receives no translational estimates. Reducing this drift for tighter spaces could be investigated by retraining with optical flow and a downward rangefinder supplied directly to the policy, rather than only through the estimator.

The comparative evaluation covers near-frontal impacts against a flat rigid wall at 3 and 5\,m/s, on one platform with motion-capture aiding. The two exploratory recoveries at 6.02 and 6.05\,m/s do not establish reliability or spatial requirements at those speeds. Oblique and glancing collisions, other surfaces and platforms, and alternative aiding modalities remain untested. All experiments also use an external position-based trigger, selected for comparability; onboard proprioceptive triggering remains to be demonstrated despite the source-independent mode-entry interface.

Finally, frame compliance affects both impact survival and the distribution of post-impact states. Joint optimization of frame stiffness and recovery control could improve recovery performance and is a further direction for study.

\section{Conclusions}
\label{sec:conclusion}

REFLEX provides a robust and autonomous recovery mode for collision-resilient soft quadrotors that runs entirely on the flight controller. In the tested stock PX4 configuration, valid external velocity measurements are rejected after high-speed wall impacts, leaving nominal control with erroneous translational feedback. REFLEX combines proprioceptive recurrent stabilization, in-flight estimator reinitialization, and consistency-gated nominal return to position control, without contact sensors or a companion computer.

In hardware experiments, REFLEX recovers 19 of 22 impacts (86\%) at $5.03\pm0.09$\,m/s, including 15 consecutive recoveries in one flight, compared with 1 of 14 (7\%) for the matched stock autopilot. A separate exploratory flight demonstrates recoveries at 6.02 and 6.05\,m/s, exceeding the previously reported velocity for autonomous wall-collision recovery by 64\%. All three unsuccessful 5\,m/s REFLEX trials fail after handback, motivating continued estimator monitoring and broader state-consistency checks. These results support integrated recovery as a fallback for autonomous flight on the tested collision-resilient platform.

\bibliographystyle{IEEEtran}
\bibliography{references}

\end{document}